%% file: mmsivae_main.tex
\documentclass[pmlr,twocolumn,10pt]{jmlr} 

\usepackage{algorithm}
\usepackage{algpseudocode}
\usepackage{booktabs}
\usepackage{multirow}
\usepackage{multicol}
\usepackage{siunitx}

\usepackage[switch]{lineno}

\theorembodyfont{\upshape}
\theoremheaderfont{\scshape}
\theorempostheader{:}
\theoremsep{\newline}

\jmlrvolume{LEAVE UNSET}
\jmlryear{2026}
\jmlrsubmitted{LEAVE UNSET}
\jmlrpublished{LEAVE UNSET}
\jmlrworkshop{Conference on Health, Inference, and Learning (CHIL) 2026} 

\title[Multimodal normative modeling with introspective variational autoencoders]{Multimodal normative modeling in Alzheimer's Disease with introspective variational autoencoders}

\author{%
\Name{Sayantan Kumar} \Email{sayantan.kumar@wustl.edu}\\
\addr Department of Computer Science and Engineering \\ Washington University in St Louis, USA
\AND
\Name{Peijie Qiu}\Email{peijie.qiu@wustl.edu}\\
\addr Malinkrodt Institute of Radiology \\Washington University in St Louis School of Medicine, USA
\AND
\Name{Aristeidis Sotiras} \Email{aristeidis.sotiras@wustl.edu}\\
\addr Malinkrodt Institute of Radiology \\ Washington University in St Louis School of Medicine
}

\begin{document}

\maketitle

\begin{abstract}
Normative modeling learns a healthy reference distribution and quantifies subject-specific deviations to capture heterogeneous disease effects. In Alzheimer’s disease (AD), multimodal neuroimaging offers complementary signals but VAE-based normative models often (i) fit the healthy reference distribution imperfectly, inflating false positives, and (ii) use posterior aggregation (e.g., PoE/MoE) that can yield weak multimodal fusion in the shared latent space. We propose \emph{mmSIVAE}, a multimodal soft-introspective variational autoencoder combined with \emph{Mixture-of-Product-of-Experts} (MOPOE) aggregation to improve reference fidelity and multimodal integration. We compute deviation scores in latent space and feature space as distances from the learned healthy distributions, and map statistically significant latent deviations to regional abnormalities for interpretability. On ADNI MRI regional volumes and amyloid PET SUVR, mmSIVAE improves reconstruction on held-out controls and produces more discriminative deviation scores for outlier detection than VAE baselines, with higher likelihood ratios and clearer separation between control and AD-spectrum cohorts. Deviation maps highlight region-level patterns aligned with established AD-related changes. More broadly, our results highlight the importance of training objectives that prioritize reference-distribution fidelity and robust multimodal posterior aggregation for normative modeling, with implications for deviation-based analysis across multimodal clinical data.
\end{abstract}

\paragraph*{Data and Code Availability}
This study uses data from the Alzheimer's Disease Neuroimaging Initiative (ADNI) which are publicly available and can be requested following ADNI Data Sharing and Publications Committee guidelines: \url{https://adni.loni.usc.edu/data-samples/access-data/}. Code scripts will be made public upon acceptance.

\paragraph*{Institutional Review Board (IRB)}
All ADNI participants provided written informed consent, and study protocols were approved by each local site's institutional review board. 


\section{Introduction}
\input{Sections/introduction}


\begin{figure*}[hbt!]
    \centering
    \includegraphics[width = 0.8\linewidth]{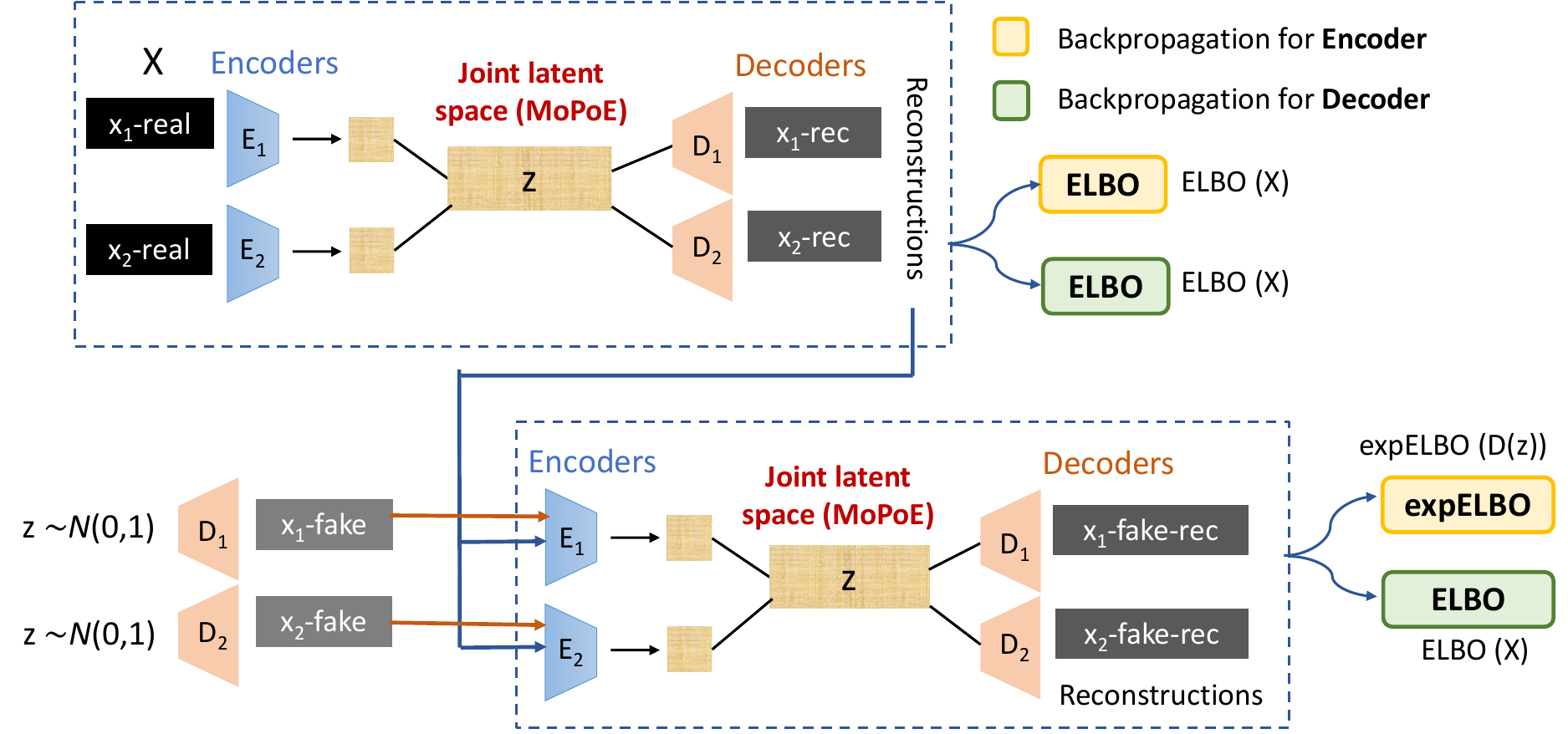}
    \vspace{-10pt}
    \caption{Training flow of mmSIVAE. The ELBO for real samples is optimized for both encoders and decoders, while the encoders also optimize the expELBO to ’push away’ generated samples from the latent space. The decoders optimize the ELBO for the generated samples to 'fool' the encoders.}
    \vspace{-15pt}
    \label{fig:mmsivae}
\end{figure*}


\section{Background and Preliminaries}
\input{Sections/background_og}

\begin{figure*}[!hbtp]
    \centering
    \includegraphics[width = 0.75\linewidth]{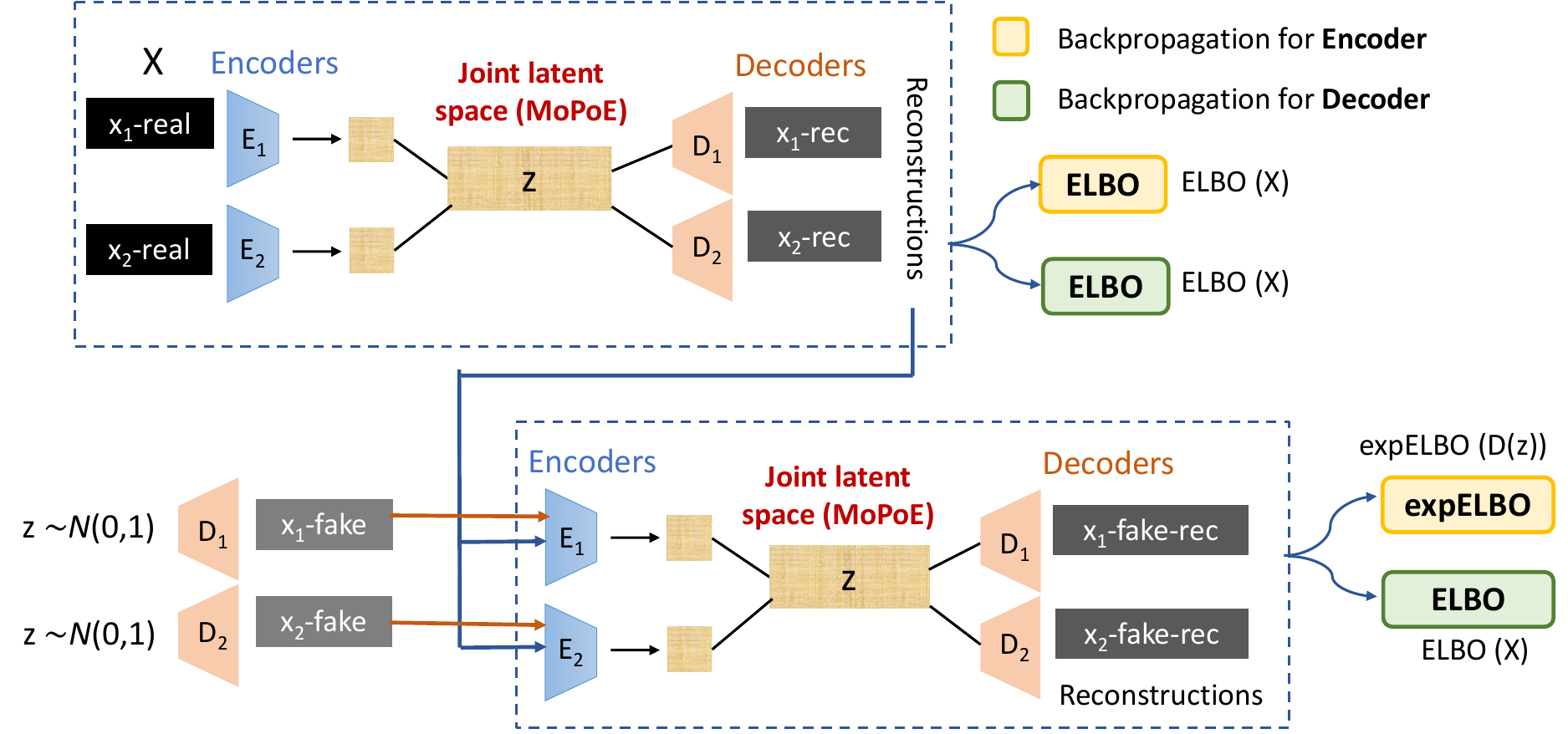}
    \vspace{-10pt}
    \caption{Training flow of mmSIVAE. The ELBO for real samples is optimized for both encoders and decoders, while the encoders also optimize the expELBO to ’push away’ generated samples from the latent space. The decoders optimize the ELBO for the generated samples to 'fool' the encoders.}
    \vspace{-10pt}
    \label{fig:mmsivae}
\end{figure*}


\section{Proposed Methodology}
\input{Sections/methods_v2}

\section{Experiments and Results}

\input{Sections/experimental_results}

\vspace{-2.5mm}
\section{Discussion}

\input{Sections/discussion_v2}


\bibliography{references}

\clearpage
\appendix

\input{Sections/appendix_mmSIVAE_proof}
\input{Sections/appendix_MOPOE}
\input{Sections/appendix_model}


\end{document}

%% file: Sections/introduction.tex
Brain disorders such as Alzheimer’s disease (AD) affect millions of people worldwide and substantially reduce quality of life for patients and their families \cite{kumar_machine_2021,richards_what_2010}. Although advances in biomarkers and therapeutics have been significant, much of the empirical literature continues to rely on case–control comparisons that emphasize group-level averages, implicitly assuming relatively homogeneous disease effects within diagnostic categories \cite{verdi2021beyond}. In AD, however, clinical presentation, underlying pathology, and rates of progression vary markedly across individuals. Quantifying \emph{individual-level} departures from expected trajectories therefore offers an opportunity to better capture disease heterogeneity, improve stratification, and support more precise hypotheses about disease mechanisms and treatment response \cite{marquand_beyond_2016,marquand_conceptualizing_2019,jack_hypothetical_2010,kumar2024analysing}.

Normative modeling addresses this need by learning the distribution of measurements in a reference population (typically cognitively unimpaired controls) and quantifying how much each individual deviates from that learned "norm" \cite{kia_neural_2019,marquand_conceptualizing_2019,verdi_revealing_2023}. Classical normative approaches (e.g., Gaussian regression, w-scores) are often univariate and can miss clinically relevant multivariate interactions across brain regions and biomarkers \cite{verdi_revealing_2023,loreto_alzheimers_2024,earnest_datadriven_2024}. Deep generative frameworks, including autoencoders and variational autoencoders (VAEs), offer a compelling alternative: they can model complex non-linear structure, embed subjects in a latent space, and provide subject-specific deviation maps by contrasting reconstructions with observed data \cite{kumar_improving_2023,lawry_aguila_multi-modal_2023}.

However, two persistent limitations reduce the reliability and interpretability of VAE-based normative modeling in AD. First, the normative paradigm depends on learning the healthy (training) distribution sufficiently well that reconstructions for controls are accurate; otherwise, large residuals may occur even for controls, inflating false positives and blurring the separation between normal and abnormal subjects. Empirically, standard, variational, and adversarial autoencoders can underperform on out-of-distribution (OOD) detection when normal and abnormal distributions overlap \cite{bercea2023aes}. Second, most deep normative models remain unimodal, even though AD is multifactorial and routinely studied with complementary neuroimaging modalities (e.g., MRI for neurodegeneration and PET for molecular pathology). Restricting to a single modality can discard information that is essential for capturing heterogeneous disease expression \cite{kumar2024analysing,lawry_aguila_conditional_2022}.

Multimodal VAEs can, in principle, address the latter limitation, but in practice they introduce an additional challenge: how to aggregate modality-specific posteriors into a joint latent representation. Common strategies include Product-of-Experts (PoE) and Mixture-of-Experts (MoE). PoE can over-weight a highly precise modality, effectively suppressing weaker modalities and yielding brittle joint inference \cite{wu_multimodal_2018}. MoE, conversely, spreads probability mass across experts and often produces a joint posterior that is no sharper than any unimodal posterior, limiting the utility of the shared latent space for deviation detection \cite{shi2019variational}. When the joint latent distribution is uninformative, subject-level deviation scores derived from that space become less sensitive and less clinically meaningful.

To address this, our work proposes a multimodal normative modeling framework designed to (i) learn the healthy distribution more faithfully and (ii) form a more informative multimodal latent representation. We introduce a \emph{multimodal soft-introspective variational autoencoder} (mmSIVAE) that extends Soft-IntroVAE (SIVAE) \cite{daniel2021soft} to the multimodal setting. SIVAE trains a VAE with an introspective (adversarial) mechanism in which the encoder is encouraged to distinguish real from generated samples, improving generation quality and OOD sensitivity \cite{daniel2021soft}. We complement this with Mixture-of-Product-of-Experts (MOPOE) aggregation to combine the strengths of PoE and MoE, yielding a shared latent space that better supports normative deviation scoring.

Our contributions can be summarized as follows:
\begin{itemize}
\item We propose \textbf{mmSIVAE}, a multimodal extension of soft-introspective VAEs for normative modeling across multiple neuroimaging modalities, and provide a theoretical analysis of the encoder-decoder game for multiple modalities
\item We adopt \textbf{MOPOE} aggregation to improve multimodal posterior inference, mitigating failure modes of PoE and MoE and producing a more informative shared latent space for downstream deviation quantification.
\item We define \textbf{multimodal normative deviation scores} in the learned shared latent space and clinically validate their sensitivity to AD stage and association with cognitive measures.
\item To improve interpretability, we \textbf{map abnormal latent deviations to regional feature-space deviations} and analyze brain regions with deviations consistent with AD pathology.
\end{itemize}

Overall, the goal is to assess whether mmSIVAE with MOPOE can (a) better capture the distribution of a healthy reference population, (b) reconstruct inputs more accurately from multimodal latent representations, and (c) yield deviation measures that are both discriminative for outlier detection and clinically interpretable in AD.

%% file: Sections/background_og.tex
Let $X = [x_1, x_2,...x_N]$ be a set of conditionally independent N modalities. The backbone of our model is a multimodal variational autoencoder (mVAE), a generative model of the form $p_\theta(x_1, x_2,...x_N, z) = p(z)\prod_{i=1}^{N}p_\theta(x_i|z)$, where z is the latent variable and $p(z)$ is the prior. mVAE optimizes the ELBO (Evidence Lower Bound) which is a combination of modality-specific likelihood distributions $p_\theta\left(x_i|z\right)$ and the KL divergence between the approximate joint posterior $q(z|X)$ and prior $p(z)$. 

\vspace{-5mm}
\begin{equation}
\begin{aligned}
\operatorname{ELBO} &= \mathbb{E}_{q(z \mid X)}\left[\sum_{x_i \in X} \log p_\theta\left(x_i|z\right)\right] \\
&-\operatorname{KL}\left[q_\phi(z|X),p(z)\right] \label{eq:mmELBO}
\end{aligned}
\end{equation}
\vspace{-5mm}

\subsection{Soft-Introspective VAE (SIVAE)}

\textbf{Introspective VAE:} Unlike the standard VAE, which optimizes a single lower bound, Introspective VAE (IntroVAE) (\cite{huang2018introvae}) incorporates an adversarial learning strategy similar to that used in Generative Adversarial Networks (GANs). In this framework, the encoder's objective is to maximize the KL divergence between the generated (fake) image and the latent variable while minimizing the KL divergence between the actual image and the latent variable. Simultaneously, the decoder seeks to challenge the encoder by minimizing the KL divergence between the generated image and the latent variable. The learning objectives for the Encoder and Decoder in IntroVAE are as follows:

\vspace{-5mm}
\begin{equation}
\begin{aligned}
& \mathcal{L}_E=E L B O(x)+\sum_{s=r, g}\left[m-K L\left(q_\phi\left(z \mid x_s\right) \| p(z)\right]^{+}\right. \\
& \mathcal{L}_D=\sum_{s=r, g}\left[K L\left(q_\phi\left(z \mid x_s\right) \| p(z)\right)\right] .
\end{aligned}
\end{equation}
\vspace{-5mm}

where $x_r$ is the reconstructed image, $xg$ is the generated image, and $m$ is the hard threshold for constraining the KL divergence.

\paragraph{Soft-Introspective VAE:} A key limitation of IntroVAE is its use of a hard threshold m to constrain the KL divergence term. Soft-Introspective VAE (SIVAE; \cite{daniel2021soft}) argue that this design significantly reduces the model's capacity and can lead to vanishing gradients. To address this issue, SIVAE proposes utilizing the entire Evidence Lower Bound (ELBO) rather than just the KL divergence, employing a soft exponential function in place of a hard threshold. The learning objective (i.e., loss function) for SIVAE is as follows:

\vspace{-5mm}
\begin{equation}
\begin{aligned}
& \mathcal{L}_{E_\phi}(x, z)=E L B O(x)-\frac{1}{\alpha} \exp \left(\alpha E L B O\left(D_\theta(z)\right)\right) \\
& \mathcal{L}_{D_\theta}(x, z)=E L B O(x)+\gamma E L B O\left(D_\theta(z)\right) \label{SIVAE}
\end{aligned}
\end{equation}
\vspace{-5mm}

\subsection{Approximating joint posterior in the latent space}

\textbf{Product-of-Experts (POE):} The approximate joint posterior $q_{PoE} (z \mid x)$ can be estimated as the Product of Experts (POE), where the experts are the unimodal approximate posteriors $q_\phi (z \mid x_i)$ and $p(z)$ and $p(z)$ is the "prior" expert.

\vspace{-3mm}
\begin{equation}
\begin{aligned}
q_{PoE}(z \mid X) &= p(z) \prod_{i = 1}^{N} q_\phi(z \mid x_i) \label{POE}
\end{aligned}
\end{equation}
\vspace{-3mm}

The product distribution required above are not in general solvable in closed form. However, if we approximate both $p(z)$ and $\Tilde{q}(z|x_i)$ as Gaussian, a product of Gaussian experts is itself Gaussian with mean $\mu = (\sum_{i} \mu_i*T_i)(\sum_{i}T_i)^{-1} $ and variance $\sigma = (\sum_{i}T_i)^{-1}$  where $\mu_i$ and $\sigma_i$ are parameters of the i-th Gaussian expert and $T_i = \sigma_i^{-1}$. 

The challenge of approximating inference distribution with POE is that if the inference distribution for a particular modality (expert) is very sharp, then the joint distribution will be dominated by it. In other words, the overconfident but mis-calibrated experts may bias the joint posterior distribution which is undesirable for learning informative latent representations between modalities.

\paragraph{Mixture-of-Experts (MOE):} Another way to approximate joint inference is to use the mixture of experts (MoE) form, where the approximate posterior of the joint is represented by the sum of the unimodal posterior distributions. Since MoE is the sum of each expert and not product, the joint posterior distribution is not dominated by experts with high precision as in PoE, but spreads its density over all individual experts. In the MoE setting, each uni-modal posterior $q_\phi(z \mid x_i)$ is evaluated with the generative model $p_\theta (X, z)$ such that the ELBO becomes:

\vspace{-5mm}
\begin{equation}
\begin{aligned}
q_{MoE}(z \mid X) &= \frac{1}{N} \sum_{i = 1}^{N} q_\phi(z \mid x_i) \\
\operatorname{ELBO}(X) &\triangleq \mathbb{E}_{q(z \mid X)}\left[\sum_{x_i \in X} \log p_\theta\left(x_i \mid z\right)\right] \\
&-\operatorname{KL}\left[q_{MoE}(z \mid X)\| p(z)\right] \\
&= \frac{1}{N}\left[\sum_{i=1}^{N}\mathbb{E}_{q(z \mid X)} \log p_\theta\left(x_i \mid z\right)\right] \\
&- \operatorname{KL}\left[ \frac{1}{N} \sum_{i = 1}^{N} q_\phi(z \mid x_i) \| p(z)\right]
\end{aligned}
\end{equation}
\vspace{-3mm}

But this approach only takes each unimodal encoding distribution separately into account during training. The aggregation of experts in MoE does not result in a distribution that is sharper than the other experts. Therefore, even if we increase the number of experts, the shared representation does not become more informative as in PoE.  Thus there is no explicit aggregation of information from multiple modalities in the latent representation for reconstruction by the decoder networks.

%% file: Sections/methods_v2.tex


In this section, we introduce multimodal soft-introspective VAE (mmSIVAE), extending SIVAE (\cite{daniel2021soft}) to the multimodal setting. First, we present the theoretical analysis of mmSIVAE, followed by modeling of the joint latent posterior using the Mixture-of-Product-of-Experts (MOPOE) technique. Next, we discuss how mmSIVAE can be adopted for multimodal normative modeling which includes model training, inference and calculation of subject-level deviations.

Following Equation \eqref{SIVAE}, the encoder and decoder loss functions for mmSIVAE can be written as:

\vspace{-3mm}
\begin{equation}
\begin{aligned}
& \mathcal{L}_{E_\phi}(X, z)=E L B O(X)-\frac{1}{\alpha} \exp \left(\alpha E L B O\left(D(z)\right)\right) \\
& \mathcal{L}_{D_\theta}(X, z)=E L B O(X)+\gamma E L B O\left(D(z)\right) \label{eq:mmSIVAE}
\end{aligned}
\end{equation}
\vspace{-3mm}

where $X=\left\{x_i \mid i^{th} \text{modality}\right\}$, and $\alpha >= 0$, $\gamma >= 0$ are hyperparameters. $ELBO(X)$ is defined in Equation \eqref{eq:mmELBO}. $ELBO(D(z))$ can be defined similarly, but with $D_\theta(z)_i = p_\theta\left(x_i \mid z\right)$ replacing x in Equation \eqref{eq:mmELBO}. Following $\operatorname{ELBO}(D(z)$ in Equation \ref{eq:mmELBO}, $\operatorname{ELBO}(D(z)$ can be written as: 

\vspace{-5mm}
\begin{equation}
\begin{aligned}
\operatorname{ELBO}(D(z)) &\triangleq \mathbb{E}_{q(z \mid X)}\left[\sum_{D_\theta(z)_i \in D} \log p_\theta\left(D_\theta(z)_i \mid z\right)\right] \\
&-\operatorname{KL}\left[q_\phi(z \mid D)\| p(z)\right]
\end{aligned}
\end{equation}
\label{ELBO(D(z)}
\vspace{-5mm}

Equation \ref{eq:mmSIVAE} represents a min-max game between the encoders and the decoders. The encoders are encouraged, via the ELBO value, to differentiate between real samples (high ELBO) and generated samples (low ELBO), while the decoders aim to generate samples that “fool” the encoders (Figure \ref{fig:mmsivae}).

\subsection{Theoretical analysis}
\label{subsec:theory}

We extend the nonparametric analysis of SIVAE \cite{daniel2021soft} to multimodal data and analyze the Nash equilibrium of the min--max game in Eq.~\eqref{SIVAE}.
We represent the encoder by the approximate joint posterior $q=q(z|X)$ and the decoder by $d=p_d(X|z)$.
Let $p_{\text{data}}(X)$ denote the data distribution and $p(z)$ a prior over $z$.
Using conditional independence of modalities and linearity of expectation, the marginal distribution of generated samples for modality $i$ is $p_d(x_i)=\mathbb{E}_{p(z)}[p_d(x_i\mid z)]$.

We write the ELBO for real samples and for generated samples as $W(X;d,q)$ and $W(D;d,q)$, respectively, where $D$ denotes the collection of modality-specific decoder outputs $D=\{D_\theta(z)_i\}$:
\begin{equation}
\begin{aligned}
W(X;d,q) &\doteq 
\mathbb{E}_{q(z \mid X)}
\Big[\sum_{x_i \in X}\log p_d(x_i \mid z)\Big]
\\ 
&-\mathrm{KL}\!\left(q(z \mid X)\,\|\,p(z)\right),\\
W(D;d,q) &\doteq \mathbb{E}_{q(z \mid X)}
\Big[\sum_{D_\theta(z)_i \in D}\log p_d(D_\theta(z)_i \mid z)\Big]
\\ 
&-\mathrm{KL}\!\left(q(z \mid D)\,\|\,p(z)\right).
\end{aligned}
\end{equation}

In a nonparametric setting (where $d$ and $q$ can be any distributions), the encoder and decoder objectives can be written as:
\begin{equation}
\begin{aligned}
\mathcal{L}_{E_\phi}(x,z)
&=
W(X;d,q)
-\alpha^{-1}\exp\!\big(\alpha\,W(D(z);d,q)\big),\\
\mathcal{L}_{D_\theta}(x,z)
&=
W(X;d,q)
+\gamma\,W(D(z);d,q).
\end{aligned}
\end{equation}
Taking expectations over real samples ($p_{\text{data}}$) and generated samples ($p_d$) yields the complete objectives:
\begin{equation}
\begin{aligned}
L_q(q,d)
&=
\mathbb{E}_{p_{\text{data}}}\!\big[W(X;d,q)\big] \\
&-\mathbb{E}_{p_d}\!\big[\alpha^{-1}\exp(\alpha W(X;d,q))\big],\\
L_d(q,d)
&=
\mathbb{E}_{p_{\text{data}}}\!\big[W(X;d,q)\big]
+\gamma\,\mathbb{E}_{p_d}\!\big[W(X;d,q)\big].
\end{aligned}
\label{complete_mmSIVAE_objective}
\end{equation}
A Nash equilibrium $(q^*,d^*)$ satisfies $L_q(q^*,d^*) \ge L_q(q,d^*)$ and $L_d(q^*,d^*) \ge L_d(q^*,d)$ for all $q,d$.

\paragraph{Lemma 1.}
If $\sum_i p_d(x_i) \le p_{\text{data}}(X)^{\frac{1}{\alpha+1}}$ for all $X$ such that $p_{\text{data}}(X)>0$, then an optimal encoder response $q^*(d)$ satisfies $q^*(d)(z|X)=p_d(z|X)$, and $W(X;q^*(d),d)=\sum_i \log p_d(x_i)$.

\paragraph{Lemma 2.}
Let $d^*$ be defined so that
$d^*\in \arg\min_d \{\mathrm{KL}(p_{\text{data}}(X)\,\|\,p_d(X))+\gamma\,H(p_d(X))\}$,
where $H(\cdot)$ is Shannon entropy. If $q^*=p_{d^*}(z\mid x)$, then $(q^*,d^*)$ is a Nash equilibrium of Eq.~\eqref{complete_mmSIVAE_objective}.

Proofs of Lemma~1 and Lemma~2 are provided in Appendix \ref{app:mmSIVAE_nash}.

\subsection{Modeling of the joint posterior -- MOPOE}
\label{subsec:mopoe}

We adopt Mixture-of-Product-of-Experts (MOPOE), a generalization of both PoE and MoE, to aggregate unimodal posteriors into a shared latent distribution.
Let $X_k$ denote a (non-empty) subset of the $N$ modalities and let $P(X)$ denote the power set.
We define:
\begin{equation}
\begin{aligned}
q_{PoE}(z \mid X_k)
&\propto \prod_{x_i \in X_k} q(z|x_i),\\
q_{MOPOE}(z|X)
&=
\frac{1}{2^N}\sum_{X_k \in P(X)} q_{PoE}(z|X_k),\\
ELBO_{MOPOE}(X)
&=
\mathbb{E}_{q(z \mid X)}
\Big[\sum_{x_i \in X} \log p_\theta(x_i|z)\Big] \\
-&\mathrm{KL}\!\left(q_{MOPOE}(z|X)\,\|\,p(z)\right).
\end{aligned}
\label{eq:mopoe_block}
\end{equation}
MOPOE can be viewed as a hierarchical aggregation: unimodal posteriors for each subset $X_k$ are first merged via PoE, and subset-level approximations are then combined via MoE.

\paragraph{Lemma 3.}
MOPOE yields a valid generalized multimodal ELBO, with PoE and MoE as special cases.

A concise proof is provided in Appendix \ref{app:mopoe_elbo}.

\subsection{Multimodal normative modeling}
\label{subsec:normative}

mmSIVAE (Fig.~1) uses separate encoders to map the two modalities to unimodal latent parameters (mean and variance), which are aggregated via MOPOE to obtain shared latent parameters (a joint latent distribution). The shared latent is then passed through modality-specific decoders to reconstruct each modality. We first train the model to characterize the healthy population cohort. We assume disease abnormality can be quantified by measuring how subjects deviate from the joint latent space \cite{lawry_aguila_multi-modal_2023,lawry_aguila_conditional_2022} or from reconstruction errors relative to healthy controls \cite{kumar_normative_2023,kumar2024analysing}. During inference, the trained model is applied to the AD cohort to estimate both latent and feature-space deviations. The detailed training procedure of mmSIVAE along with the pseudo code (Algorithm \ref{alg:training_algo}) and complete set of hyperparameters are presented in Appendix \ref{app:model_training}.

\input{Figures_tables/healthy_reconstruction}


\subsection{Patient-specific deviations -- latent and feature space}

\textbf{Mahalanobis distance.}
To quantify how much each subject deviates from the latent distribution of healthy controls, we compute the Mahalanobis distance \cite{lawry2023multi}, which accounts for correlations between latent vectors:
\begin{equation}
D_{ml}
=
\sqrt{(z_j-\mu(z_{norm}))^\top
\Sigma(z_{norm})^{-1}
(z_j-\mu(z_{norm}))}.
\label{d_ml}
\end{equation}
Here $z_j \equiv q(z_j|X_j)$ is a sample from the joint posterior for subject $j$, and $\mu(z_{norm})$ and $\Sigma(z_{norm})$ are the mean and covariance of latent positions for healthy controls.
We also derive a multivariate feature-space deviation index based on Mahalanobis distance that quantifies how reconstruction errors deviate from the reconstruction-error distribution of controls:
\begin{equation}
D_{mf}
=
\sqrt{(d_j-\mu(d_{norm}))^\top
\Sigma(d_{norm})^{-1}
(d_j-\mu(d_{norm}))}.
\label{d_mf}
\end{equation}
Here $d_j=[d^1_j,\ldots,d^i_j,\ldots,d^R_j]$ is the mean squared reconstruction error between original and reconstructed input for subject $j$ at brain region $i\in\{1,\ldots,R\}$, and $\mu(d_{norm})$ and $\Sigma(d_{norm})$ denote the mean and covariance of reconstruction errors in healthy controls.

\textbf{Z-score latent and feature deviations.}
To identify latent dimensions and brain regions associated with abnormal deviations, we compute latent and feature-space $Z$-scores:
\begin{equation}
\begin{aligned}
Z_{ml} &= \frac{z_{ij}-\mu(z_{ij}^{norm})}{\sigma(z_{ij}^{norm})},
Z_{mf} &= \frac{d_{ij}-\mu(d_{ij}^{norm})}{\sigma(d_{ij}^{norm})}.
\end{aligned}
\label{z_ml_z_mf}
\end{equation}
Here $z_{ij}$ and $d_{ij}$ are the latent value and reconstruction error for test subject $j$ at position/region $i$, and $\mu(\cdot)$ and $\sigma(\cdot)$ are computed from healthy controls.

%% file: Figures_tables/healthy_reconstruction.tex
\begin{figure*}[!htbp]
    \centering
    \begin{minipage}{0.48\textwidth}
        \centering
        \includegraphics[width=\linewidth]{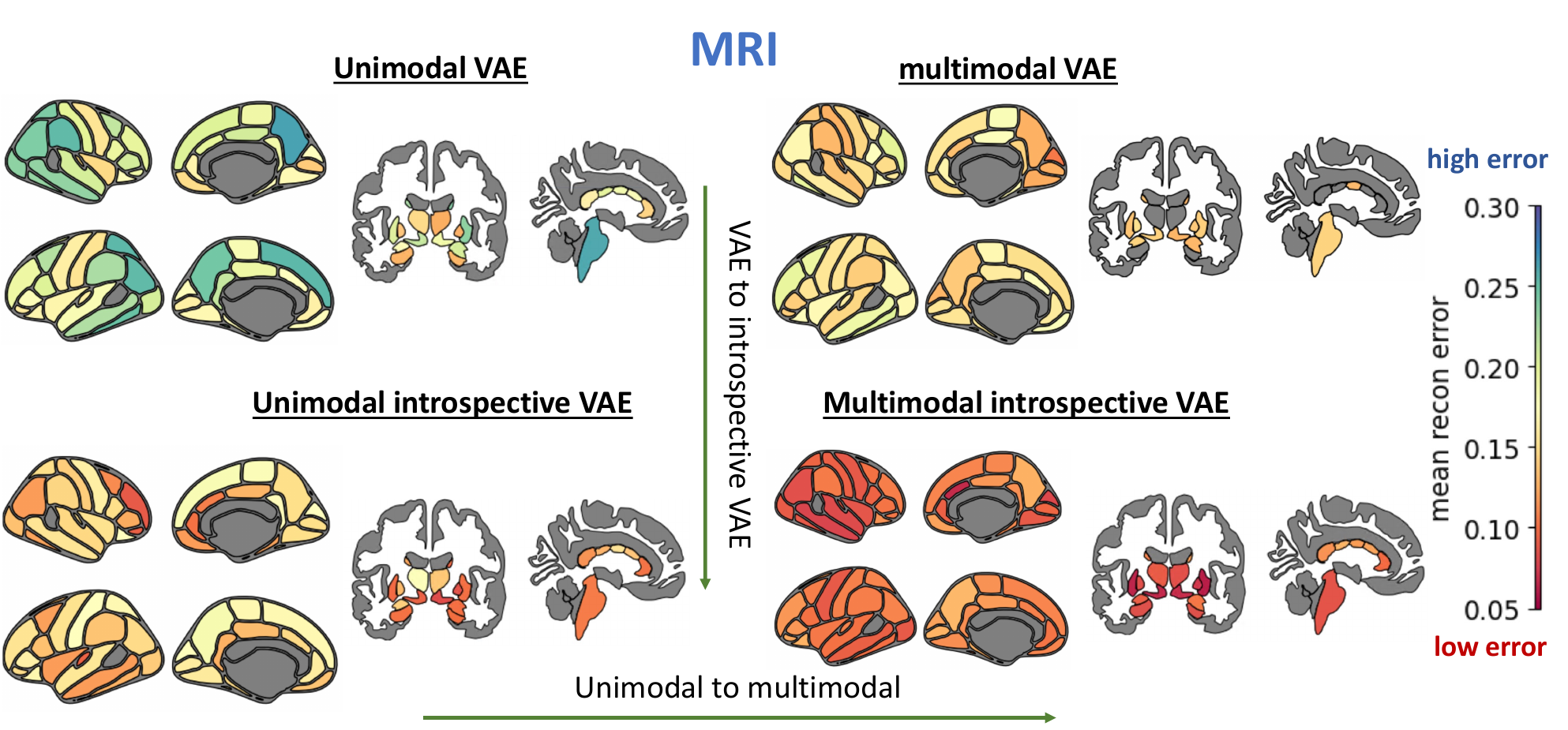}
    \end{minipage}\hfill
    \begin{minipage}{0.48\textwidth}
        \centering
        \includegraphics[width=\linewidth]{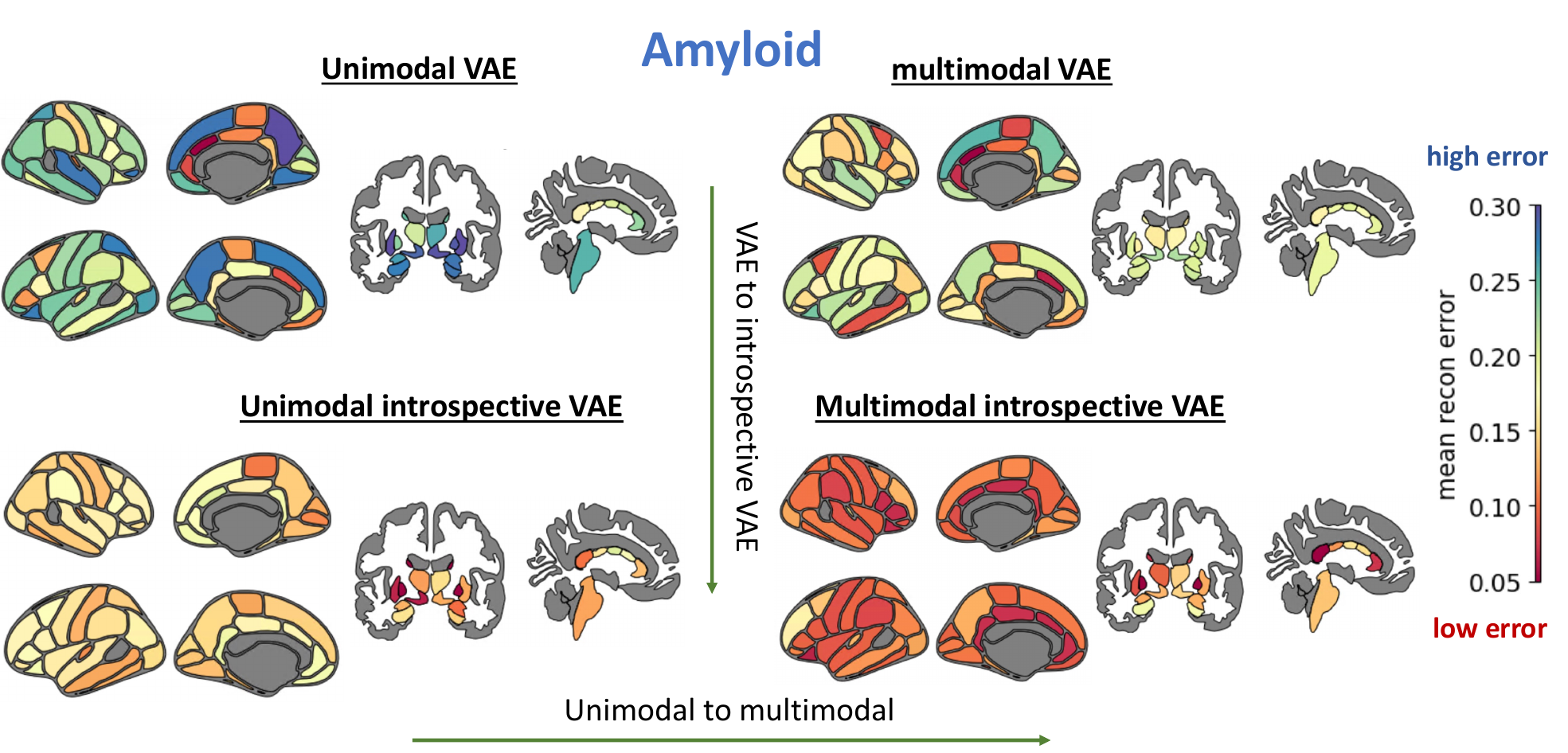}
    \end{minipage}
    \caption{Average reconstruction error when reconstructing MRI volumes (a) and amyloid SUVR (b) for our proposed mmSIVAE and baselines (SIVAE, mmVAE and unimodal VAE). The brain atlases showing the reconstruction errors for both MRI volumes and amyloid SUVR were visualized using the ggseg package in Python.}
    \label{fig:recon_comparison_Aim3}
    \vspace{-10pt}
\end{figure*}

%% file: Sections/experimental_results.tex

\subsection{Experimental setup}

\paragraph*{Data and feature preprocessing}
For training, we selected 248 cognitively unimpaired (healthy control) subjects from the Alzheimer’s Disease Neuroimaging (ADNI) dataset \cite{mueller2005ways} with Clinical Dementia Rating (CDR) = 0 and no amyloid pathology. We used regional brain volumes extracted from T1-weighted MRI scans and regional Standardized Uptake Value Ratio (SUVR) values extracted from AV45 Amyloid PET scans as the two input modalities for our model (Figure 1). Both brain volumes and SUVR values were extracted from 66 cortical (Desikan-Killiany atlas) and 24 subcortical regions. For model evaluation, we used 48 healthy controls for a separate holdout cohort and a disease cohort of 726 amyloid positive AD spectrum (ADS) individuals across the following AD stages: preclinical stage with no symptoms $(CDR = 0, A+) (N = 305)$, (b) $CDR = 0.5 (N = 236)$ and (c) $CDR >= 1 (N = 185)$.  

Each brain ROI was normalised by removing the mean and dividing by the standard deviation of the healthy control cohort brain regions. We conditioned our model on the age and sex of patients, represented as one-hot encoding vectors, to remove the effects of covariates from the MRI and PET features.

\paragraph{Baselines and implementation details}

\paragraph{Baselines:} Our proposed mmSIVAE was compared with the following baselines as follows: (i) Unimodal Soft-Instrospective VAE (SIVAE) \cite{daniel2021soft}(Multimodal VAE (Aim 2B; mmVAE) \cite{kumar_normative_2023}, (ii) unimodal VAE \cite{pinaya_using_2021} and \textbf{state-of-the-art multimodal VAE models}: mmJSD \cite{sutter2020multimodal}, JMVAE \cite{suzuki2016joint} and MVTCAE \cite{hwang2021multi}. For multimodal methods (mmSIVAE and mmVAE), model performance using different aggregation techniques (POE, MOE and MOPOE) were compared. Unimodal methods (SIVAE and VAE) had a single encoder and decoder and used either a single modality (MRI/amyloid) or both modalities as a single concatenated input. Models mmJSD, JMVAE and MVTCAE were implemented using the open-source Multi-view-AE python package \cite{aguila2023multi}.


All models were implemented in Pytorch and trained using Adam optimizer with hyperparameters as follows: epochs = 500, learning rate = $10^{-5}$, batch size = $64$ and latent dimensions in the range [5,10,15,20]. The encoder and decoder networks have 2 fully-connected layers of sizes ${64, 32}$ and ${32, 64}$ respectively.

\begin{table*}[!t]
\caption{Likelihood ratio corresponding to outliers derived from $D_{ml}$ (\textbf{A}) and $D_{mf}$ (\textbf{B}) for different latent dimensions $d = 5, 10, 15$ and $20$. Higher values indicate better detection of outlier AD patients.}
\label{tab:sig_ratio_both}
\centering
\begin{minipage}[t]{0.48\textwidth}
    \centering
\input{Figures_tables/latent_sig_ratio_table}
\end{minipage}\hfill
\begin{minipage}[t]{0.48\textwidth}
    \centering
\input{Figures_tables/feature_sig_ratio_table}
\end{minipage}
\end{table*}

\subsection{Results: reconstruction on healthy controls}
\label{subsec:results_recon}


Our first objective was to evaluate whether mmSIVAE achieved better reconstruction performance for cognitively unimpaired healthy controls compared to a standard vanilla VAE. Furthermore, we examined whether multimodal methods, which integrate information from multiple modalities in the latent space, demonstrated improved reconstruction accuracy compared to their unimodal counterparts. To address this, we visualized the average reconstruction errors for each cortical and subcortical region across our proposed mmSIVAE and baseline models, including SIVAE, multimodal VAE, and vanilla VAE (Figure \ref{fig:recon_comparison_Aim3}). For a fair comparison, both mmSIVAE and mmVAE employed the MOPOE technique to aggregate multimodal information within the latent space. In contrast, the unimodal methods, SIVAE and vanilla VAE, utilized a single encoder and decoder, with both modalities (MRI and amyloid) concatenated as a single unimodal input. 

Across all brain regions, mmSIVAE demonstrated the lowest reconstruction error for both modalities when compared to the baseline models (Figure \ref{fig:recon_comparison_Aim3}). Introspective VAE methods, such as mmSIVAE and SIVAE, outperformed vanilla VAE approaches (mmVAE and VAE) in reconstruction accuracy. Additionally, multimodal methods (mmSIVAE and mmVAE), which integrate information from multiple modalities, exhibited lower reconstruction errors than unimodal VAEs that treated multiple modalities as a single unimodal input (Figure \ref{fig:recon_comparison_Aim3}).

\begin{figure*}[!thbp]
    \centering
    \vspace{-10pt}
    \includegraphics[width = 0.75\linewidth]{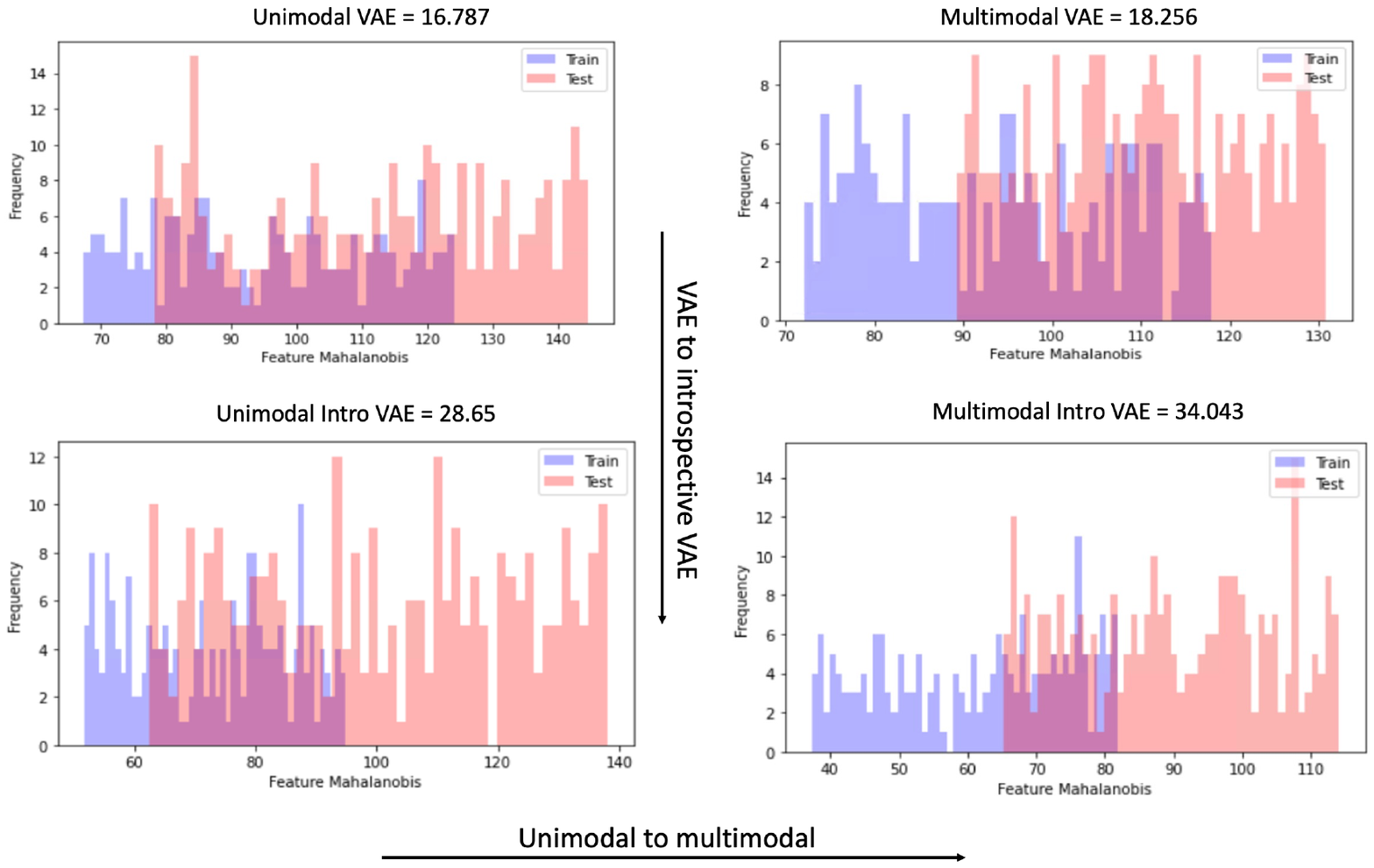}
    \vspace{-10pt}
    \caption{Histograms showing the distribution of feature mahalanobis deviations $D_{mf}$ (Eq \ref{d_mf}) for the test (AD patients) and a holdout healthy control cohort. The values in the caption of each subfigure indicate the Earth mover's distance between the 2 distributions train and test. Higher distance indicate lesser overlap/better separation between the healthy and disease cohorts.}
    \vspace{-10pt}
    \label{fig:feature_mahalanobis_histogram}
\end{figure*}
\subsection{Results: outlier detection performance}
\label{subsec:results_ood}



The latent mahalanobis deviations $D_{ml}$ (Eq \ref{d_ml}) and the feature mahalanobis deviations $D_{mf}$ (Eq \ref{d_mf}) were estimated independently for the ADS cohort and separate holdout cohort of healthy controls with no cognitive impairment. For both deviation metrics, subjects with statistically significant ($p<0.001$) deviations \cite{tabachnick2013using} were labeled as abnormal/outliers. A good normative model is supposed to correctly identify disease subjects as outliers and healthy individuals within the normative distribution. Following previous work, we used the positive likelihood ratio to assess how well our mmSIVAE model can detect outliers.

\vspace{-5mm}
\begin{equation*}
\text{likelihood ratio} = \frac{\text{TPR}}{\text{FPR}}
= \frac{\frac{N_{disease}(outliers)}{N_{disease}}}{\frac{N_{controls}(outliers)}{N_{controls}}}
\end{equation*}
\vspace{-5mm}

\paragraph{Deviations in the latent space} Latent deviations ($D_{ml}$) consistently achieved higher likelihood ratios compared to feature space deviations ($D_{mf}$), as shown in Table \ref{tab:sig_ratio_both}A. This demonstrates that calculating deviations within the multimodal joint latent space enhances normative performance and improves outlier detection compared to regional space deviations. Among the models, mmSIVAE with MOPOE latent space aggregation achieved the highest likelihood ratios, outperforming all baselines. Introspective VAE methods (mmSIVAE and SIVAE) exhibited superior normative performance compared to vanilla VAE-based methods (mmVAE and unimodal VAE). Furthermore, in both mmSIVAE and mmVAE, MOPOE aggregation proved to be more effective for outlier detection than POE or MOE, validating its selection as the preferred aggregation strategy by taking advantage of the strengths of both approaches. In general, increasing the number of latent dimensions to 15 or 20 resulted in higher likelihood ratios, further improving performance.

A similar trend is observed for likelihood ratios generated by feature-space Mahalanobis deviations ($D_{mf}$), as shown in Table \ref{tab:sig_ratio_both}B. Our proposed mmSIVAE achieved the greatest separation between control and disease cohorts, evidenced by the highest Earth Mover's Distance (34.043) between the $D_{mf}$ distributions of the two cohorts (Figure \ref{fig:feature_mahalanobis_histogram}). Introspective VAE-based methods (mmSIVAE and SIVAE) demonstrated more pronounced separation (reduced overlap and greater distance) between control and disease cohorts compared to VAE-based methods (mmVAE and VAE), as illustrated in Figure \ref{fig:feature_mahalanobis_histogram}.

\input{Figures_tables/interpretability}

\subsection{Clinical interpretability}
\label{subsec:cognition_interpretability}



\paragraph{Mapping from latent to feature deviations}

Ideally, all latent dimensions can be used to reconstruct the input data and quantify feature-space deviations. The latent dimensions with statistically significant ($p<0.05$) mean absolute Z-scores $Z_{ml}$ (Eq \ref{z_ml_z_mf}) indicate the latent dimensions which show deviation between control and disease cohorts. This can provide an interpretation of how latent space deviations can be mapped to deviations in the feature-space. We passed these selected latent vectors through the decoders setting the remaining latent dimensions and covariates to be 0 such that the reconstructions and feature deviations $Z_{mf}$ (Eq \ref{z_ml_z_mf}) reflect only the information encoded in the selected latent vectors. From Table \ref{tab:sig_ratio_both}A, we observed mmSIVAE with MOPOE latent space aggregation had the maximum likelihood ratio for d = 15 (number of latent dimensions). We identified 5 out of 15 latent dimensions (6,7,8,14 and 15) whose mean absolute $Z_{ml} > 1.96$ ($p < 0.05$) and used them for generating the feature-space deviations $Z_{mf}$ (Figure \ref{fig:interpretability}A).


\paragraph{Brain regions associated with AD abnormality}

We visualized the pairwise differences in the magnitude of abnormal deviations ($Z_{mf}$) in each region between amyloid negative CU individuals and disease groups along the ADS: (i) $CDR = 0$(preclinical AD), (ii) $CDR = 0.5$ (very mild dementia), and (iii) $CDR \geq 1$ (mild or more severe dementia). Our aim was to validate the derived regional abnormal deviations by examining whether these deviations showed increased group differences across progressive CDR stages. We quantified group differences using Cohen’s d-statistic effect size, calculated separately for each modality. A higher effect size when comparing regional MRI volumes indicated lower gray matter volume (more atrophy). Similarly, a higher effect size when comparing amyloid or tau uptake indicated elevated SUVR uptake (higher amyloid and tau loads) compared to the amyloid-negative CU group.


Region-level group differences in MRI atrophy were most evident within the temporal, parietal and hippocampal regions which is consistent with the observations in existing literature \cite{leech2014role,noh2014anatomical} (Figure 2B). Higher group differences in amyloid loading were mostly observed in the accumbens, precuneus, frontal and temporal regions, which are sensitive to amyloid pathology accumulation \cite{levitis2022differentiating}  (Figure \ref{fig:interpretability}B). 

%% file: Figures_tables/latent_sig_ratio_table.tex
\centering
\footnotesize
\setlength{\tabcolsep}{4pt}
\begin{tabular}{l c c c c}
\toprule
\textbf{Method} & \textbf{$d=5$} & \textbf{$d=10$} & \textbf{$d=15$} & \textbf{$d=20$} \\
\midrule

\textbf{mmSIVAE} & & & & \\
\quad MOPOE & 7.21 & 7.85 & \textbf{10.2} & 9.34 \\
\quad POE   & 7.05 & 6.82 & 7.86 & 8.10 \\
\quad MOE   & 7.21 & 7.85 & 9.21 & 8.56 \\
\addlinespace

\textbf{mmVAE} & & & & \\
\quad MOPOE & 6.73 & \textbf{8.10} & 5.72 & 4.56 \\
\quad MOE   & 6.26 & 7.57 & 5.45 & 5.68 \\
\quad POE   & 6.21 & 7.95 & 5.68 & 4.89 \\
\addlinespace

\textbf{Unimodal baselines} & & & & \\
\quad SIVAE (MRI+Amyloid) & 4.91 & 5.86 & 7.43 & 6.90 \\
\quad SIVAE (MRI)     & 7.22 & 6.85 & 8.56 & 8.25 \\
\quad SIVAE (Amyloid)     & 7.25 & 6.82 & 8.41 & 8.83 \\
\quad mmJSD           & 6.31 & 7.21 & 4.66 & 5.25 \\
\quad JMVAE           & 5.68 & 7.85 & 5.10 & 4.51 \\
\quad MVTCAE          & 6.34 & 6.83 & 5.56 & 3.82 \\

\bottomrule
\end{tabular}

%% file: Figures_tables/feature_sig_ratio_table.tex
\centering
\footnotesize
\setlength{\tabcolsep}{4pt}
\begin{tabular}{l c c c c}
\toprule
\textbf{Method} & \textbf{$d = 5$} & \textbf{$d = 10$} & \textbf{$d = 15$} & \textbf{$d = 20$} \\
\midrule

\textbf{mmSIVAE} & & & & \\
\quad MOPOE & 2.42 & 2.76 & \textbf{3.29} & \textbf{3.34} \\
\quad POE   & \textbf{2.57} & 2.68 & 2.85 & 3.09 \\
\quad MOE   & 2.45 & \textbf{2.77} & 2.55 & 2.94 \\
\addlinespace

\textbf{mmVAE} & & & & \\
\quad MOPOE & 1.85 & \textbf{2.52} & 2.16 & 1.76 \\
\quad MOE   & 1.67 & 1.92 & \textbf{2.22} & 1.32 \\
\quad POE   & 1.37 & 1.53 & 1.77 & \textbf{1.68} \\
\addlinespace

\textbf{Unimodal baselines} & & & & \\
\quad SIVAE (MRI+Amy) & 1.96 & 2.04 & 2.62 & 2.85 \\
\quad SIVAE (MRI)     & 2.21 & 2.45 & 2.72 & 3.05 \\
\quad SIVAE (Amy)     & 2.32 & 2.57 & 2.75 & 2.91 \\
\quad mmJSD           & 2.21 & 1.81 & 1.40 & 1.25 \\
\quad JMVAE           & 1.68 & 2.33 & 2.10 & 1.75 \\
\quad MVTCAE          & 1.35 & 1.68 & 1.52 & 1.21 \\
\bottomrule
\end{tabular}

%% file: Figures_tables/interpretability.tex
\begin{figure*}[!thbp]
    \centering
    \begin{minipage}{0.45\textwidth}
        \centering
        \includegraphics[width=\linewidth]{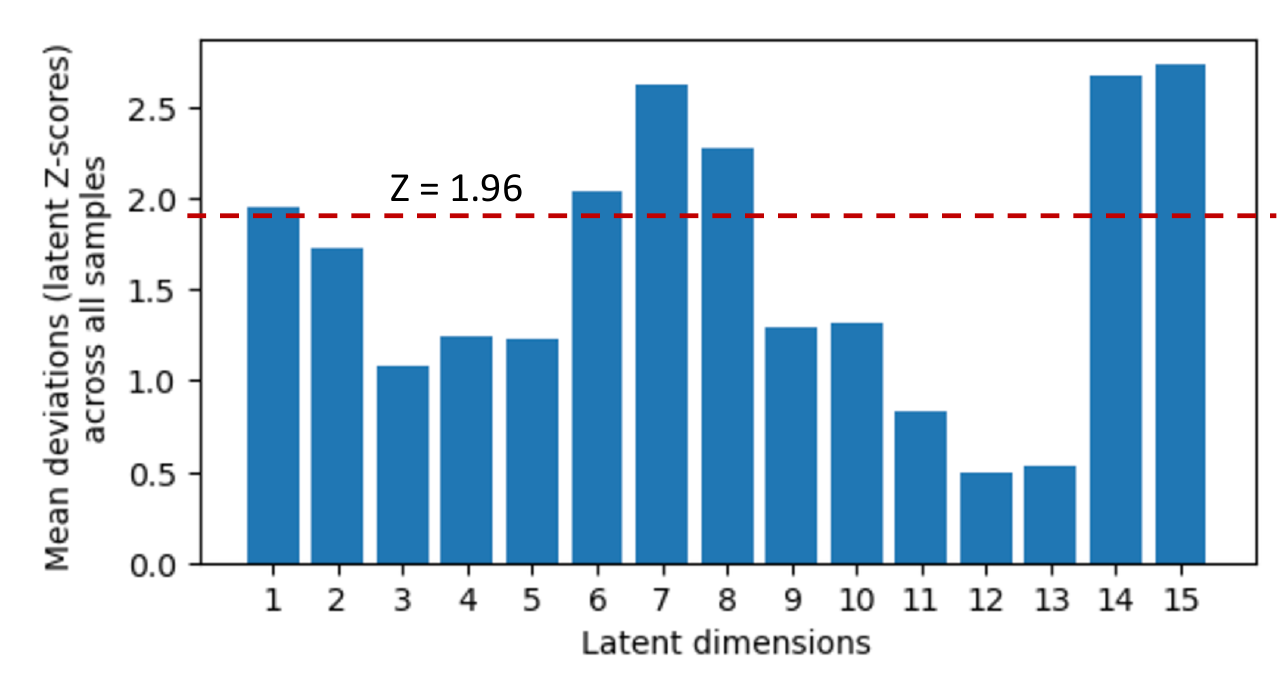}
    \end{minipage}\hfill
    \begin{minipage}{0.52\textwidth}
        \centering
        \includegraphics[width=\linewidth]{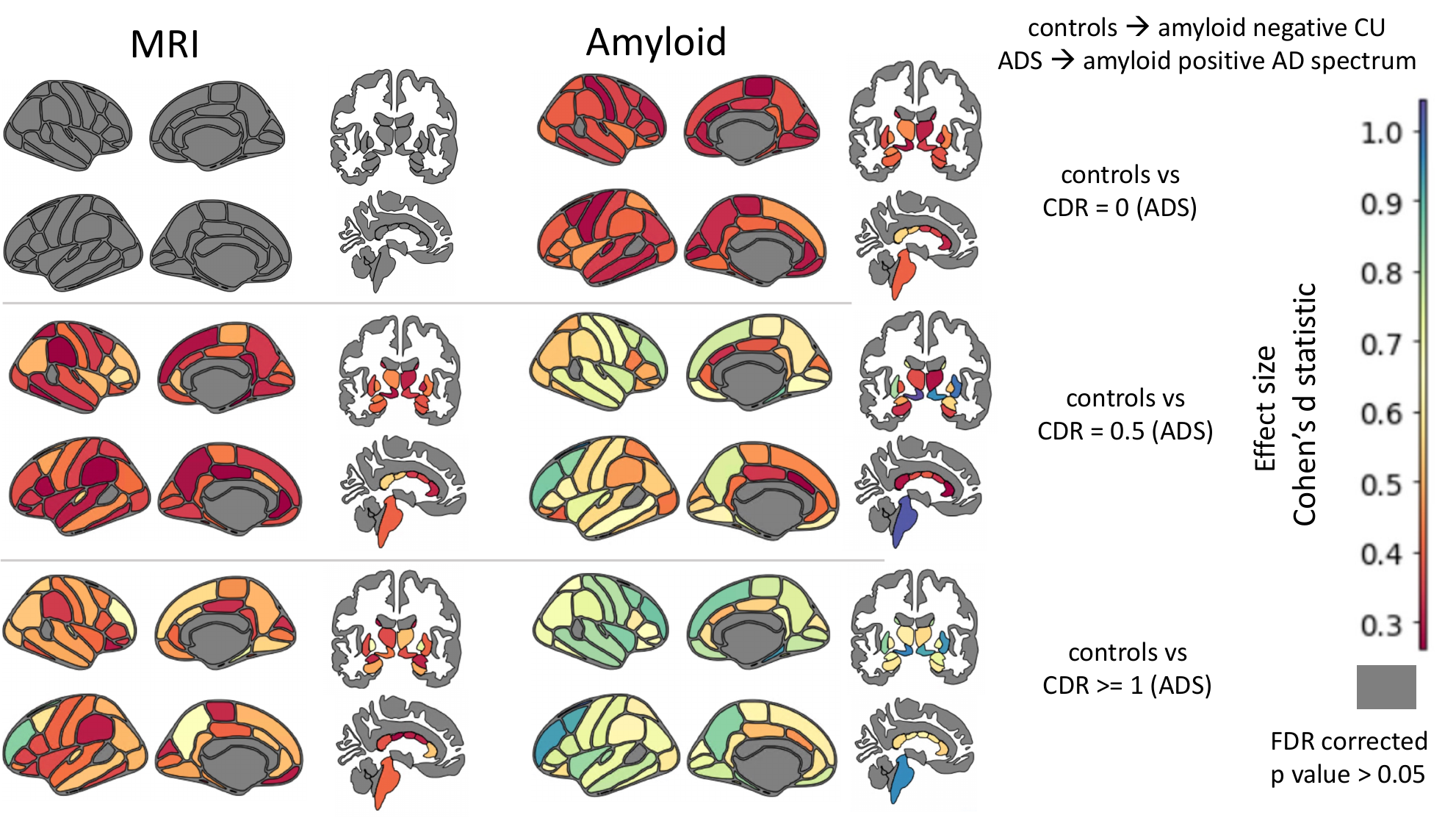}
    \end{minipage}
    \caption{\textbf{A}.Five latent dimensions (6, 7, 8, 14, and 15) out of 15 show statistically significant deviations, defined by mean absolute $Z_{ml} > 1.96$ ($p<0.05$). The dotted red line denotes the significance threshold ($Z=1.96$); latent dimensions exceeding this threshold are used for mapping to feature-space deviations. \textbf{B}. Brain atlas maps (Desikan–Killiany atlas for 66 cortical regions and Aseg atlas for 24 subcortical regions) illustrate pairwise group differences in regional deviation magnitude between control and disease groups. Colors indicate effect size (Cohen’s $d$), with $d=0.2$, $0.5$, and $0.8$ corresponding to small, medium, and large effects, respectively. Gray regions indicate no statistically significant differences after FDR correction.}
    \label{fig:interpretability}
    \vspace{-10pt}
\end{figure*}

%% file: Sections/discussion_v2.tex
Our results show that mmSIVAE---combining soft-introspective training with robust multimodal posterior aggregation---improves reference-distribution fidelity (Fig.~\ref{fig:recon_comparison_Aim3}), strengthens deviation-based separation between healthy controls and AD-spectrum cohorts (Table~\ref{tab:sig_ratio_both}A; Fig.~\ref{fig:feature_mahalanobis_histogram}), and yields clinically meaningful deviation measures with interpretable regional patterns (Fig.~\ref{fig:interpretability}). Below we summarize the main take-home messages and how they address gaps in multimodal normative modeling.

\paragraph{Benefits of soft-introspective training over vanilla VAE.}
A central challenge in normative modeling is accurately capturing the reference (healthy) distribution: when reconstruction quality is weak on held-out controls, reconstruction-based deviation scores can inflate false positives and reduce downstream separability. Relative to standard VAE training, the introspective objective explicitly pressures the model to improve fidelity on real samples while discouraging trivial solutions that make generated samples appear ``too normal,'' which helps sharpen the learned reference manifold. Empirically, this is reflected in improved reconstructions on healthy controls (Fig.~\ref{fig:recon_comparison_Aim3}) and stronger separation in deviation scores both in the latent and feature space (Table~\ref{tab:sig_ratio_both}A,B; Fig.~\ref{fig:feature_mahalanobis_histogram}). Methodologically, this suggests that for deviation-based pipelines, optimizing only the vanilla ELBO may be insufficient: training objectives that prioritize reference fidelity can have first-order impact on downstream outlier detection.

\paragraph{Aggregating multimodal information in the latent space.}
How the joint posterior $q(z\mid X)$ is constructed is not a minor implementation detail: it determines whether the shared latent representation is informative for deviation scoring. PoE aggregation can yield sharp posteriors but may be dominated by a highly precise or miscalibrated modality, suppressing complementary signal and producing brittle inference \cite{wu_multimodal_2018}. MoE aggregation improves robustness by averaging experts, yet can under-fuse modalities, often yielding a joint posterior that is no sharper than unimodal posteriors \cite{shi2019variational}. MOPOE addresses this trade-off by mixing PoE posteriors across modality subsets, retaining sharpness while reducing single-modality domination. Consistent with this motivation, we observe stronger likelihood-ratio separation under MOPOE in the latent deviations (Table~\ref{tab:sig_ratio_both}A), indicating that improved multimodal fusion translates directly into more useful deviation scores.

\paragraph{Clinical insights and generalizability.}
In heterogeneous disorders such as AD, subject-level deviations can capture variation that is not well summarized by coarse staging alone. Regional deviation maps provide a transparent view of how abnormalities manifest across MRI- and PET-derived features (Fig.~\ref{fig:interpretability}B). Importantly, the clinical take-away is not tied to ADNI-specific labels: the broader implication is that reliable multimodal normative models can produce quantitative phenotypes for stratification, hypothesis generation, and studying links between biomarkers and behavior. More generally, the combination of (i) training for reference fidelity and (ii) robust multimodal posterior aggregation is likely to be beneficial in other healthcare settings where modalities have different noise levels, missingness patterns, or clinical relevance.

\paragraph{Limitations and future directions.}
This study is evaluated on ADNI, and generalization may be affected by site/scanner effects, demographic differences, and selection bias typical of research cohorts; validating stability under explicit domain shifts and across external cohorts is an important next step. Our deviation scores $D_{ml}$ and $D_{mf}$ rely on Gaussian modeling of reference embeddings and reconstruction errors, and Mahalanobis distances can be sensitive to covariance estimation, especially at higher latent dimensionalities; regularized covariance estimators and uncertainty-aware deviation intervals could improve robustness. While our interpretability pipeline promotes sparsity by selecting significant latent dimensions and mapping them back to region-level effects, thresholds (e.g., $|z|>1.96$) and multiple-comparison controls influence which effects are highlighted; systematic sensitivity analyses would strengthen interpretability claims. Future work should also extend the framework to additional modalities and longitudinal trajectories, and study principled handling of modality missingness to preserve calibrated multimodal inference in real-world clinical workflows.

%% file: Sections/appendix_mmSIVAE_proof.tex
\section{Nash equilibrium -- mmSIVAE}
\label{app:mmSIVAE_nash}

We denote the ELBO for real samples by $W(X; d, q)$ and the ELBO for generated samples by $W(D; d, q)$:
\begin{equation}
\begin{aligned}
W(X; d, q)
&\doteq
\mathbb{E}_{q(z \mid X)}
\Big[\sum_{x_i \in X} \log p_d(x_i \mid z)\Big] \\
&-\mathrm{KL}\!\left(q(z \mid X)\,\|\,p(z)\right),\\
W(D; d, q)
&\doteq
\mathbb{E}_{q(z \mid X)}
\Big[\sum_{D_\theta(z)_i \in D}\log p_d(D_\theta(z)_i \mid z)\Big] \\
&- \mathrm{KL}\!\left(q(z \mid D)\,\|\,p(z)\right).
\end{aligned}
\end{equation}
Here, the encoder is represented by the approximate joint posterior $q=q(z\mid X)$ and the decoder is represented by $d=p_d(X\mid z)$, with prior $p(z)$.
Using conditional independence across modalities and linearity of expectation, the marginal distribution of generated samples for modality $i$ is
$p_d(x_i)=\mathbb{E}_{p(z)}[p_d(x_i \mid z)]$.

\noindent By the Radon--Nikodym theorem, we have the equalities:
\begin{equation}
\begin{aligned}
\mathbb{E}_{z \sim p(z)}
\Big[\exp\!\big(\alpha W(D; d, q)\big)\Big] 
&=\mathbb{E}_{X \sim p_d}\Big[\exp\!\big(\alpha W(X; d, q)\big)\Big],\\
\mathbb{E}_{z \sim p(z)}\big[W(D; d, q)\big] 
&=\mathbb{E}_{X \sim p_d}\big[W(X; d, q)\big].
\end{aligned}
\end{equation}

\noindent Moreover, the ELBO satisfies
\begin{equation}
\begin{aligned}
W(X; d, q)
&=
\sum_{x_i \in X}\log p_d(x_i) \\
&-\mathrm{KL}\!\left(q(z \mid X)\,\|\,p_d(z \mid X)\right)\\
&\leq
\sum_{x_i \in X}\log p_d(x_i),
\end{aligned}
\end{equation}
where $p_d(X)=\prod_{x_i \in X}p_d(x_i)$ and $\sum_{x_i \in X}\log p_d(x_i)=\log p_d(X)$.

\subsection*{Nonparametric objectives}

We consider a nonparametric setting where $d$ and $q$ can be any distributions.
For $z \sim p(z)$, let $D(z)$ be a multimodal sample from $p_d(X \mid z)$.
The encoder and decoder objectives are:
\begin{equation}
\begin{aligned}
\mathcal{L}_{E_\phi}(X, z)
&=
W(X; q, d)
-
\frac{1}{\alpha}
\exp\!\big(\alpha W(D(z); q, d)\big),\\
\mathcal{L}_{D_\theta}(X, z)
&=
W(X; q, d)
+
\gamma\,W(D(z); q, d).
\end{aligned}
\end{equation}
The complete multimodal Soft-IntroVAE objectives take expectations over real samples ($p_{\text{data}}$) and generated samples ($p_d$):
\begin{equation}
\begin{aligned}
L_q(q, d)
&=
\mathbb{E}_{p_{\text{data}}}\!\big[W(X; q, d)\big]
-
\mathbb{E}_{p_d}\!\Big[\alpha^{-1}\exp\!\big(\alpha W(X; q, d)\big)\Big],\\
L_d(q, d)
&=
\mathbb{E}_{p_{\text{data}}}\!\big[W(X; q, d)\big]
+
\gamma\,\mathbb{E}_{p_d}\!\big[W(X; q, d)\big].
\end{aligned}
\label{eq:complete_mmSIVAE_objective}
\end{equation}

A Nash equilibrium point $(q^*,d^*)$ satisfies
\[
L_q(q^*, d^*) \geq L_q(q, d^*) \quad \text{and} \quad
L_d(q^*, d^*) \geq L_d(q^*, d) \quad \forall\, q,d.
\]
Given $d$, let $q^*(d)$ satisfy $L_q(q^*(d), d) \geq L_q(q, d)$ \\ for all $q$.

\subsection*{Lemma 1 and proof}

\noindent\textbf{Lemma 1.}
If $\sum_{i} p_d(x_i) \leq p_{\text{data}}(X)^{\frac{1}{\alpha+1}}$ for all $X$ such that $p_{\text{data}}(X) > 0$, then $q^*(d)$ satisfies
$q^*(d)(z\mid X) = p_d(z\mid X)$, and
$W(X; q^*(d), d) = \sum_i \log p_d(x_i)$.

\noindent\textbf{Proof.}
Using the ELBO identity,
\begin{equation}
\begin{aligned}
L_q(q, d)
&=
\mathbb{E}_{p_{\text{data}}}
\Big[
\sum_i\log p_d(x_i)\Big] \\
&-\Big[\mathrm{KL}\!\left(q(z \mid X)\,\|\,p_d(z \mid X)\right)
\Big]\\
&\quad
-\frac{1}{\alpha}\,
\mathbb{E}_{p_d}\Big[
\exp\!\Big(
\alpha \sum_i\log p_d(x_i) \\
&-\alpha \mathrm{KL}\!\left(q(z \mid X)\,\|\,p_d(z \mid X)\right)
\Big)\Big]\\
&=
\sum_X p_{\text{data}}(X)
\Big(\sum_i \log p_d(x_i) \\
&-\mathrm{KL}\!\left(q(z \mid X)\,\|\,p_d(z \mid X)\right)\Big)\\
&\quad
-\frac{1}{\alpha}\,
\prod_i p_d^{\alpha+1}(x_i)\,
\exp\!\Big(
-\alpha\,\mathrm{KL}\!\left(q(z \mid X)\,\|\,p_d(z \mid X)\right)
\Big).
\end{aligned}
\end{equation}

Fix an $X$ with $p_{\text{data}}(X)>0$ and consider maximizing the following term in $L_q$ over $q(\cdot\mid X)$:
\begin{equation}
\begin{aligned}
&p_{\text{data}}(X)
\Big[
\sum_i \log p_d(x_i)
-
\mathrm{KL}\!\left(q(z \mid X)\,\|\,p_d(z \mid X)\right)
\Big]\\
&\quad
-\frac{1}{\alpha}\,
\prod_i p_d^{\alpha+1}(x_i)\,
\exp\!\Big(
-\alpha\,\mathrm{KL}\!\left(q(z \mid X)\,\|\,p_d(z \mid X)\right)
\Big).
\end{aligned}
\end{equation}
Let $y \doteq -\mathrm{KL}(q(z \mid X)\,\|\,p_d(z \mid X)) \le 0$ and
$a \doteq \frac{\prod_i p_d^{\alpha+1}(x_i)}{p_{\text{data}}(X)}$.
Then the dependence on $q$ is through
$g(y)=y-\frac{a}{\alpha}\exp(\alpha y)$, whose maximizer is $y=-\frac{1}{\alpha}\log(a)$ when $a>1$ and $y=0$ when $a\le 1$.
Under the condition of the lemma, $a\le 1$, so the maximum is attained at $y=0$, i.e.,
$\mathrm{KL}(q(z \mid X)\,\|\,p_d(z \mid X))=0$, implying $q^*(d)(z\mid X)=p_d(z\mid X)$.
Substituting into $W(X; d, q)$ yields $W(X; q^*(d), d)=\sum_{x_i\in X}\log p_d(x_i)$.
\hfill$\square$

\subsection*{Lemma 2 and proof}

\noindent \textbf{Lemma 2.}
Let $d^*$ satisfy
\begin{equation}
d^*
\in
\underset{d}{\arg\min}
\Big\{
\mathrm{KL}\!\left(p_{\text{data}}(X)\,\|\,p_d(X)\right)
+
\gamma H\!\left(p_d(X)\right)
\Big\},
\end{equation}
where $H(\cdot)$ is the Shannon entropy of generated samples.
Let $q^* = p_{d^*}(z \mid X)$. Then $(q^*, d^*)$ is a Nash equilibrium of the min--max game in Eq.~\eqref{eq:complete_mmSIVAE_objective}.

\noindent \textbf{Proof.}
From Lemma 1, for any decoder $d$ we have $q^*(d)(z\mid X)=p_d(z\mid X)$.
Using $W(X; q^*(d), d)=\log p_d(X)-\mathrm{KL}(q^*(d)(z \mid X)\,\|\,p_d(z \mid X))$,
\begin{equation}
\begin{aligned}
L_d(q^*(d), d)
&=\mathbb{E}_{p_{\text{data}}}\big[W(X; q^*(d), d)\big] \\
&+ \gamma\,\mathbb{E}_{p_d}\big[W(X; q^*(d), d)\big]\\
&=\mathbb{E}_{p_{\text{data}}}[\log p_d(X)] \\
&-\mathbb{E}_{p_{\text{data}}}\!\Big[
\mathrm{KL}\!\left(q^*(d)(z \mid X)\,\|\,p_d(z \mid X)\right)\Big]\\
&\quad \\
&+\gamma\,\mathbb{E}_{p_d}[\log p_d(X)] \\
&-\gamma\,\mathbb{E}_{p_d}\!\Big[
\mathrm{KL}\!\left(q^*(d)(z \mid X)\,\|\,p_d(z \mid X)\right)\Big].
\end{aligned}
\end{equation}
Dropping the non-positive KL terms gives the upper bound
\begin{equation}
L_d(q^*(d), d)
\le
\mathbb{E}_{p_{\text{data}}}[\log p_d(X)]
+\gamma\,\mathbb{E}_{p_d}[\log p_d(X)].
\end{equation}
Noting $\mathbb{E}_{p_d}[\log p_d(X)]= -H(p_d(X))$ and
$\mathbb{E}_{p_{\text{data}}}[\log p_d(X)]
=
-\mathrm{KL}(p_{\text{data}}(X)\,\|\,p_d(X)) + \mathbb{E}_{p_{\text{data}}}[\log p_{\text{data}}(X)]$,
we obtain
\begin{equation}
\begin{aligned}
L_d(q^*(d), d)
&\le
-\mathrm{KL}\!\left(p_{\text{data}}(X)\,\|\,p_d(X)\right)
&-\gamma H\!\left(p_d(X)\right) \\
&+\mathbb{E}_{p_{\text{data}}}[\log p_{\text{data}}(X)].
\end{aligned}
\end{equation}
The final expectation is constant in $d$, hence maximizing $L_d(q^*(d), d)$ over $d$ is equivalent to minimizing
$\mathrm{KL}(p_{\text{data}}(X)\,\|\,p_d(X))+\gamma H(p_d(X))$, which is achieved by $d^*$ by definition.
Finally, with $q^* = p_{d^*}(z\mid X)$, Lemma 1 implies $q^*$ is a best response to $d^*$, and $d^*$ is a best response to $q^*$, so $(q^*, d^*)$ is a Nash equilibrium.
\hfill$\square$

%% file: Sections/appendix_MOPOE.tex
\section{MOPOE - generalized ELBO}
\label{app:mopoe_elbo}

\paragraph{Lemma 3: MoPoE generates a generalized ELBO with PoE and MoE being special cases.}

To prove this lemma, first we need to show that $\operatorname{ELBO}_{MoPoE}(X)$ is a valid multimodal ELBO, i.e. $\log p_\theta(x) \geq \operatorname{ELBO}_{MoPoE}(X)$. Next we will show that MOPOE is a generalized case of both POE and MOE.

\subsection*{Proof : MOPOE is a valid multimodal ELBO}

In order to prove that $\operatorname{ELBO}_{MoPoE}(X)$ is a valid multimodal ELBO, we need to show that $\log p_\theta(X) \geq \operatorname{ELBO}_{MoPoE}(X)$. Note that while taking all possible subsets $X_k \in P(X)$, we denote each subset by $k$ for better readability.

\vspace{-5mm}
\begin{equation}
\begin{aligned}
\log p_\theta(X) &=  \operatorname{KL}\left[q_{\phi}(z \mid X) \| p(z \mid X)\right] \\
&+ \mathbb{E}_{q_\phi(z \mid X)}\left[\log \frac{p(z, X)}{q_\phi(z \mid X)}\right] \\
&= \operatorname{KL}\left[\frac{1}{2^M} \sum_{k} \Tilde{q}(z \mid X_k) \| p(z \mid X)\right] \\
&+ \mathbb{E}_{\frac{1}{2^M} \sum_{k} \Tilde{q}(z \mid X_k)}\left[\log \frac{p(z, X)}{\frac{1}{2^M} \sum_{k} \Tilde{q}(z \mid X_k)}\right] \\
& \geq  \mathbb{E}_{\frac{1}{2^M} \sum_{k} \Tilde{q}(z \mid X_k)}\left[\log \frac{p(z, X)}{\frac{1}{2^M} \sum_{k} \Tilde{q}(z \mid X_k)}\right] \\
&= \mathbb{E}_{\frac{1}{2^M} \sum_{k} \Tilde{q}(z \mid X_k)}\left[\log \frac{p(X \mid z)p(z)}{\frac{1}{2^M} \sum_{k} \Tilde{q}(z \mid X_k)}\right] \\
&= \mathbb{E}_{\Tilde{q}(z \mid X)} \left[\log p_\theta(X \mid z)\right] \\
&- \mathbb{E}_{\frac{1}{2^M} \sum_{k} \Tilde{q}(z \mid X_k)}\left[\log \frac{\frac{1}{2^M} \sum_{k} \Tilde{q}(z \mid X_k)}{p(z)}\right] \\
&= \mathbb{E}_{\Tilde{q}(z \mid X)} \left[\log p_\theta(X \mid z)\right] \\
&- \operatorname{KL}\left[\frac{1}{2^M} \sum_{X_k \in P(X)} q_\phi(z \mid X_k)\| p(z)\right] \\
&= ELBO_{MoPoE}
\end{aligned}
\end{equation}
\vspace{-5mm}


\subsection*{Proof : MOPOE is a generalized version of POE and MOE}

\noindent MoPoE generates a generalized ELBO with PoE and MoE being special cases. The MVAE architecture proposed by Wu and Goodman, 2018 only takes into account the full subset, i.e., the PoE of all data types. Trivially, this is a MoE with only a single component as follows:

\vspace{-5mm}
\begin{equation}
\begin{aligned}
L_{PoE} &= \mathbb{E}_{q(z \mid X)}\left[\log p_\theta(x \mid z)\right]-K L(q(z \mid X) \| p(z)) \\ 
q_\phi(z \mid X) &\propto \prod_{j = 1}^{N} q_{\phi_j}(z \mid x_j) \\ &= PoE (\{q_{\phi_j}(z \mid x_j)\}_{j = 1}^{N}) \\ &= \sum_{k = 1}^{1} oE (\{q_{\phi_j}(z \mid x_j)\}_{j = 1}^{N})
\end{aligned}
\end{equation}
\vspace{-5mm}

\noindent As the PoE of a single expert is just the expert itself, the MMVAE model (Shi et al., 2019) is the special case of MoPoE which takes only into account the N unimodal subsets as follows:

\vspace{-5mm}
\begin{equation}
\begin{aligned}
\operatorname{ELBO}(X) &= \mathbb{E}_{q(z \mid X)}\left[\log p_\theta(X \mid z)\right] \\
&- \operatorname{KL}\left[ \frac{1}{N} \sum_{j = 1}^{N} q_{\phi_j}(z \mid x_j) \| p(z)\right] \\
\text{wirh} q_\phi(z \mid X) &\propto \frac{1}{N}\sum_{j = 1}^{N} q_{\phi_j}(z \mid x_j) \\
&= \frac{1}{N} PoE (q_{\phi_j}(z \mid x_j))
\end{aligned}
\end{equation}
\vspace{-2.5mm}

\noindent $L_{MoE}$ is equivalent to a $L_{MoPoE}$ of the N unimodal posterior approximations $q_{\phi_j}(z\mid x_j )$ for $j = 1, . . . , N$.

\noindent Therefore, the proposed MoPoE-VAE is a generalized formulation of the MVAE and MMVAE, which accounts for all subsets of modalities. The identified special cases offer a new perspective on the strengths and weaknesses of prior work: previous models focus on a specific subset of posteriors, which might lead to a decreased performance on the remaining subsets. In particular, the MVAE should perform best when all modalities are present, whereas the MMVAE should be most suitable when only a single modality is observed.

%% file: Sections/appendix_model.tex
\section{Model training and hyperparameters}
\label{app:model_training}

\input{training_algo}

Algorithm 1 shows the detailed training procedure of mmSIVAE. Expanding mmSIVAE's objective function, which is minimized from Equation \ref{eq:mmSIVAE} with the complete set of hyperparameters.


\begin{equation}
\begin{aligned}
\mathcal{L}_{E_\phi}(X, z)
&=
s \cdot \Big(
\beta_{\text{rec}} \mathcal{L}_r(X)
+ \beta_{\text{kl}} \mathrm{KL}(X)
\Big) \\
&\quad
+ \frac{1}{2}
\exp\!\Big(
-2s \cdot \big(
\beta_{\text{rec}} \mathcal{L}_r(D_\theta(z)) \\
&+ \beta_{\text{neg}} \mathrm{KL}(D_\theta(z))
\big)
\Big).
\end{aligned}
\label{mmSIVAE_complete_encoder}
\end{equation}

\begin{equation}
\begin{aligned}
\mathcal{L}_{D_\theta}(X, z)
&=
s \cdot \beta_{\text{rec}} \mathcal{L}_r(X) \\
&\quad
+ s \cdot \Big(
\beta_{\text{kl}} \mathrm{KL}(D_\theta(z)) \\
&+ \gamma_r \cdot \beta_{\text{rec}} \mathcal{L}_r(D_\theta(z))
\Big).
\end{aligned}
\label{mmSIVAE_complete_decoder}
\end{equation}

where $\mathcal{L}_r(X) = \mathbb{E}_{q(z \mid X)}\left[\sum_{x_i \in X} \log p_\theta\left(x_i \mid z\right)\right]$ denotes the summation of reconstruction error of real samples X across the 2 modalities. $KL(X) = \operatorname{KL}\left[q(z \mid X)\| p(z)\right]$ denotes the KL divergence between the joint posterior and prior $p(z)$.

$\mathcal{L}_r(D_\theta(z)) = \mathbb{E}_{q(z \mid X)}\left[\sum_{D_\theta(z)_i \in D}\log p_d\left(D_\theta(z)_i \mid z\right)\right]$ denote the summation of the reconstruction error of the generated samples $D(z)$ across the 2 modalities. $KL(D(z)) = KL(q(z \mid D) \| p(z)$ denote the KL divergence between the joint posterior of generated samples $D(z)$ and the prior $p(z)$. 

The hyperparameters $\beta_\text{rec}$ and
$\beta_\text{kl}$ control the balance between inference and sampling quality respectively. When $\beta_\text{rec} > \beta_\text{kl}$, the optimization is focused on good reconstructions, which may lead to less variability in the generated samples, as latent posteriors are allowed to be very different from the prior. When $\beta_\text{rec} < \beta_\text{kl}$, there will be more varied samples, but reconstruction quality will degrade.

Each ELBO term in Equations \ref{mmSIVAE_complete_encoder} and \ref{mmSIVAE_complete_decoder} can be considered as an instance of $\beta_\text{VAE}$ and can have different $\beta_\text{rec}$ and $\beta_\text{kl}$ parameters. However, we set them all to be the same, except for the ELBO inside the exponent in Equation \ref{mmSIVAE_complete_encoder}. For this term, $\beta_\text{kl}$ controls the repulsion force of the posterior for generated samples from the prior. We denote this specific parameter as $\beta_\text{neg}$. 

As reported in the SIVAE paper \cite{daniel2021soft}, the decoder tries to minimize the reconstruction error for generated data, which may slow down convergence, as at the beginning of the optimization the generated samples are of low quality. $\gamma_r$ is a hyperparameter that multiplies only the reconstruction term of the generated data in the ELBO term of the decoder in Equation \ref{mmSIVAE_complete_decoder}. 

Finally, we use a scaling constant s to balance between the ELBO and the expELBO terms in the loss, and we set s to be the inverse of the input dimensions \cite{daniel2021soft}. This scaling constant prevents the expELBO from vanishing for high-dimensional input in Equation \ref{mmSIVAE_complete_encoder}. 

\paragraph{Model hyperparameters:} The range of the search was [0.05, 1.0] for $\beta_\text{kl}$ and $\beta_\text{rec}$ and [$\beta_\text{kl}$, 5$\beta_\text{kl}$] for $\beta_\text{neg}$. Following a hyperparameter tuning by grid search, the final values selected are $\beta_\text{kl} = \beta_\text{kl} = 1$ and $\beta_\text{neg} = 10$. As followed in the SIVAE paper, we set constant $\gamma_r = 10^{-8}$ and s = 1/number of features = 1/90. 


%% file: training_algo.tex
\begin{algorithm*}
\caption{Training multimodal soft-introVAE (mmSIVAE)}
\label{alg:training_algo}
\begin{algorithmic}[1]
\Require $\beta_{\text{rec}}, \beta_{kl}, \beta_{\text{neg}}, \gamma_r$
\State $\phi_E, \theta_D \gets \text{Initialize network parameters}$
\State $s \gets 1 / \text{input dim}$ \Comment{Scaling constant}
\While{not converged}
    \State $X=\left\{x_i \mid i^{th} \text{modality}\right\}$ \Comment{Random mini-batch of multimodal data from dataset}
    \State $Z \gets MOPOE(E_i(x_i))$ \Comment{Encode and MOPOE aggregation in the latent space}
    \State $Z_f \gets \text{Samples from prior } N(0, I)$
    \Procedure{UpdateEncoder}{$\phi_E$}
        \State $X_r \gets \{D_i(Z)\}, \, X_f \gets \{D_i(Z_f)\}$ \Comment{Modality-specific decoders}
        \State $Z_{rf} \gets MOPOE(E(X_r^i)), \, Z_{ff} \gets MOPOE(E(X_f^i))$
        \State $X_{rf} \gets \{D_i(Z_{rf})\}, \, X_{ff} \gets \{D_i(Z_{ff})\}$
        \State $\text{ELBO} \gets -s \cdot \text{ELBO}(\beta_{\text{rec}}, \beta_{kl}, X, Z)$
        \State $\text{ELBO}_r \gets \text{ELBO}(\beta_{\text{rec}}, \beta_{\text{neg}}, X_r, X_f, Z_{rf})$
        \State $\text{ELBO}_f \gets \text{ELBO}(\beta_{\text{rec}}, \beta_{\text{neg}}, X_f, X_{ff}, Z_{ff})$
        \State $\text{expELBO}_r \gets 0.5 \cdot \exp(2s \cdot \text{ELBO}_r)$
        \State $\text{expELBO}_f \gets 0.5 \cdot \exp(2s \cdot \text{ELBO}_f)$
        \State $L_E \gets \text{ELBO} - 0.5 \cdot (\text{expELBO}_r + \text{expELBO}_f)$
        \State $\phi_E \gets \phi_E + \eta \nabla_{\phi_E}(L_E)$ \Comment{Adam update (ascend)}
    \EndProcedure
    \Procedure{UpdateDecoder}{$\theta_D$}
        \State $X_r \gets \{D_i(Z)\}, \, X_f \gets \{D_i(Z_f)\}$ \Comment{Modality-specific decoders}
        \State $Z_{rf} \gets MOPOE(E(X_r^i)), \, Z_{ff} \gets MOPOE(E(X_f^i))$
        \State $X_{rf} \gets \text{sg}\{D_i(Z_{rf})\}, \, X_{ff} \gets \text{sg}\{D_i(Z_{ff})\}$ \Comment{stop gradient}
        \State $\text{ELBO} \gets \beta_{\text{rec}} L_{\text{rec}}(X, X_r)$
        \State $\text{ELBO}_r \gets \text{ELBO}(\gamma_r \cdot \beta_{\text{rec}}, \beta_{kl}, X_r, X_f, Z_{rf})$
        \State $\text{ELBO}_f \gets \text{ELBO}(\gamma_r \cdot \beta_{\text{rec}}, \beta_{kl}, X_f, X_{ff}, Z_{ff})$
        \State $L_D \gets s \cdot (\text{ELBO} + 0.5 \cdot (\text{ELBO}_r + \text{ELBO}_f))$
        \State $\theta_D \gets \theta_D + \eta \nabla_{\theta_D}(L_D)$ \Comment{Adam update (ascend)}
    \EndProcedure

\Function{ELBO}{$\beta_{\text{rec}}, \beta_{kl}, X, Z$}
    \State $\text{ELBO} \gets -1 \cdot (\beta_{\text{rec}} L_{\text{rec}}(X, X_r) + \beta_{kl} KL(Z))$
    \State \Return $\text{ELBO}$
\EndFunction
\end{algorithmic}
\end{algorithm*}